\documentclass[journal]{IEEEtran}
\usepackage{amsfonts,graphicx}
\usepackage[hyphens]{url}
\usepackage{amsmath, mathabx, amsthm}
\usepackage{array}
\usepackage[implicit=false,hidelinks]{hyperref}
\usepackage[dvipsnames]{xcolor}
\usepackage{cite}

\usepackage{tikz}
\usetikzlibrary{matrix}

\newcommand{\specialcell}[2][c]{%
  \begin{tabular}[#1]{@{}c@{}}#2\end{tabular}}

\newcommand{\argmin}{\mathop{\mathrm{arg\,min}}\limits}
\newcommand{\norm}[1]{\left\lVert#1\right\rVert}
\newcommand{\matr}[1]{\mathbf{#1}}
\newcommand{\vect}[1]{\mathbf{#1}}

\begin{document}
\title{A Fourier Disparity Layer representation for Light Fields}

\author{Mikael~Le Pendu,
        Christine~Guillemot,
        and~Aljosa~Smolic

\thanks{This project has been supported in part by the Science Foundation Ireland (SFI) under the Grant Number 15/RP/2776 and in part by the EU H2020 Research and Innovation Programme under grant agreement No 694122 (ERC advanced grant CLIM).}%
}

\maketitle

\begin{abstract}
In this paper, we present a new Light Field representation for efficient Light Field processing and rendering called Fourier Disparity Layers (FDL).
The proposed FDL representation samples the Light Field in the depth (or equivalently the disparity) dimension by decomposing the scene as a discrete sum of layers. The layers can be constructed from various types of Light Field inputs including a set of sub-aperture images, a focal stack, or even a combination of both. From our derivations in the Fourier domain, the layers are simply obtained by a regularized least square regression performed independently at each spatial frequency, which is efficiently parallelized in a GPU implementation.
Our model is also used to derive a gradient descent based calibration step that estimates the input view positions and an optimal set of disparity values required for the layer construction.
Once the layers are known, they can be simply shifted and filtered to produce different viewpoints of the scene while controlling the focus and simulating a camera aperture of arbitrary shape and size. Our implementation in the Fourier domain allows real time Light Field rendering.
Finally, direct applications such as view interpolation or extrapolation and denoising are presented and evaluated.


\end{abstract}


\section{Introduction}

Light Fields are commonly represented as 4 dimensional functions with 2 spatial and 2 angular dimensions \cite{Levoy1996, Gortler1996}. They can be seen as 2D arrays of images (called sub-aperture images), each having an unlimited depth of field, and differing from their neighbour images only by a slight shift of the view angle. The sampling in the angular dimensions is key in Light Field imaging \cite{PlenopticSampling}. In particular, densely sampled Light Fields make it possible to directly render images with shallow depth of field while controlling the focus depth.
Such rendering, often referred to as Light Field refocusing, does not require knowledge of the scene's geometry. It is usually performed either by shifting and averaging the sub-aperture images \cite{Ng2005a} or by selecting a 2D slice in the 4D Fourier domain \cite{Ng2005b}.
However, a dense angular sampling comes at the expense of very high requirements in terms of capture, storage, processing power and memory.
A too sparse angular sampling, on the other hand, does not allow for a smooth transition between viewpoints and causes angular aliasing in the refocused images, characterized by sharp structures in the out of focus regions. The importance of a dense angular sampling is clearly shown by the vast literature on viewpoint interpolation. Several approaches exist including depth image based rendering techniques \cite{Georgiev2006, WAN13PAMI, Pujades2014, Chaurasia2013, Zhang2015, Soft3DReconstruction}, deep learning methods either exploiting a depth map estimation \cite{Kalantari2016, Flynn2015} or not \cite{Wu2017, Yoon2015}, and approaches leveraging sparsity priors of the Light Field data in a transformed domain \cite{Shi2014, Takahashi2017, Vaghar2017, Vaghar2018}.
However, although viewpoint interpolation greatly simplifies the capture of dense Light Fields, it also increases the amount of data to store and process for the final rendering application.

Alternatively, Light Fields can be represented as a focal stack, that is, a set of shallow depth of field images (e.g. photos taken with a wide aperture) with different focusing depths.
This representation has the advantage of allowing an unlimited angular density with few images because the sampling is performed on the depth dimension instead of 2 angular dimensions. However, for rendering tasks such as simulating a different camera aperture size or shape, or a change of viewpoint, the common approach is to first convert the focal stack into the 4D representation. For instance, Levin and Durand \cite{Levin2010} retrieve sub-aperture images by the deconvolution of shifted and averaged focal stack images. A similar deconvolution technique is used in \cite{Kodama2013} to first synthesize the 4D Light Field from a focal stack in order to render images with arbitrary aperture shapes. More recent methods have also been proposed to reconstruct the 4D Light Field from a focal stack either in the spatial domain using depth from focus \cite{Mousnier2015}, or via optimization in the Fourier domain \cite{Perez2016, Alonso2016}.

\begin{figure}
\centerline{\includegraphics[width=\linewidth]{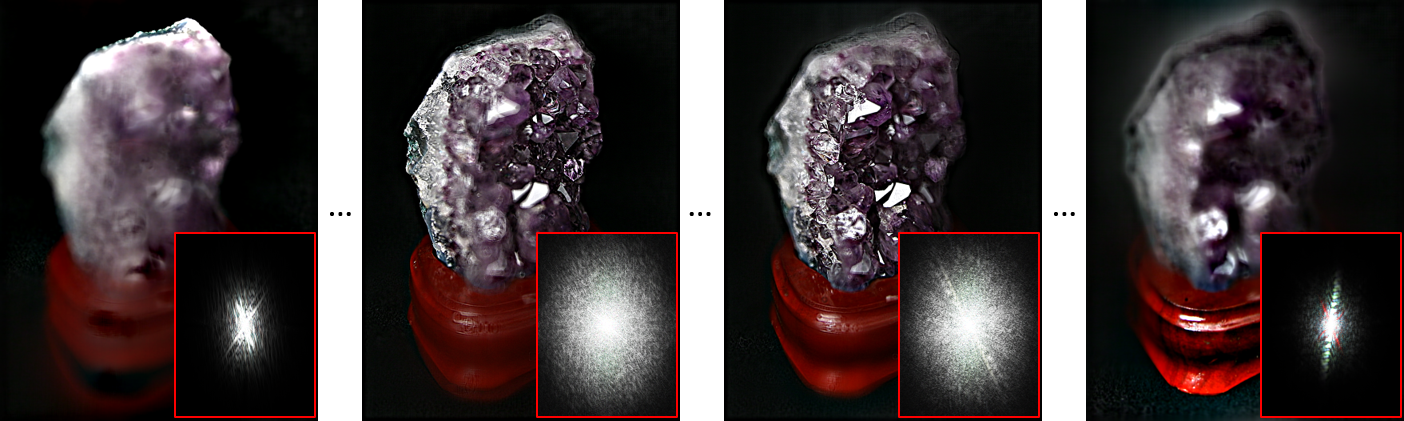}}
\vspace{-6pt}
\caption{
Fourier Disparity Layer representation. For the visualization, the layers are shown in the spatial domain (i.e. after inverse Fourier Transform). The magnitude spectrum of each layer is also shown in the red boxes.
Note that the FDL representation differs visually from a focal stack: the out-of-focus regions tend to disappear while the regions in focus have enhanced contrasts.}
\label{fig:Illustration}
\vspace{-15pt}
\end{figure}

In this paper, we propose the Fourier Disparity Layer (FDL) representation illustrated in Fig. \ref{fig:Illustration}. It can be easily constructed either from a 4D Light Field, a focal stack, or even a hybrid Light Field combining sub-aperture and wide aperture images with varying parameters (i.e. focusing depth, aperture, point of view).
Given the 2D Fourier transform of each input image with their parameters (i.e. angular coordinates, aperture, focus), the FDL model is constructed using linear optimization. For each frequency component, a linear least squares problem is solved to determine the corresponding Fourier coefficients of the different layers, each layer being associated to a given disparity value.
The layers can then be directly used for real time rendering. For instance, sub-aperture images are obtained by shifting the layers proportionally to their associated disparity value and by averaging them. This is directly implemented in the Fourier domain as a simple linear combination with frequency-dependent coefficients. More general rendering with arbitrary point of view, aperture shape and size, and focusing depth is performed with the same computational complexity without the need to first reconstruct the 4D Light Field.

In the case where the input is a set of sub-aperture images, we propose a gradient descent based calibration method to determine their angular coordinates as well as the optimal set of disparity values. The formulation of the optimization problems for the calibration and the layer construction are closely related. Nevertheless, we define two regularization schemes with different properties to better suit each situation.


Additionally, we demonstrate the effectiveness of our approach for several direct applications.
First, when the input is a sparse set of sub-aperture images, view interpolation and extrapolation is obtained by constructing the FDL representation and by rendering views at intermediate angular coordinates with an infinitely small aperture. In a second application, the same viewpoints as the input are rendered to produce a denoised result. For this use case, we present a possible extension of the model where the shift applied to each layer is not constrained to be proportional to the angular coordinates of the view to reconstruct. The relaxed model allows a more accurate representation of occlusions and non-Lambertian effects in the scene.

Since the computational complexity is a key aspect motivating the need for a new Light Field representation, our implementation makes efficient use of the GPU at every step of the processing chain (i.e. calibration, layer construction, rendering). The proposed algorithms are built upon simple linear algebra operations performed independently at each frequency component, which makes our approach particularly suitable for GPU parallelization.

In summary, the contributions are:
\begin{itemize}
\item Definition of the Fourier Disparity Layer representation and its construction from other Light Field representations (e.g. sub-aperture images, focal stack, combination of focal stack images and sub-aperture images).
\item Calibration method jointly determining the input view positions and disparity values of the layers.
\item Fast and advanced Light Field rendering from the FDL representation with simultaneous control over the viewpoint, aperture size, aperture shape and focusing depth.
\item Analysis of other application scenarios: view interpolation and denoising.
\end{itemize}

\section{Related work}

Related Light Field representations have been used in the design of several Light Field displays \cite{Wetzstein2011, Wetzstein2012, Takahashi2018}. These displays reproduce the Light Field using a stack of light attenuating LCD layers placed in front of a backlight. Thanks to the distance separating the LCD panels in the display, the image perceived depends on the observer's position and is proportional to the product of the layers. This layer representation has similarities with the one presented in this paper, and it can be constructed either from the Light Field views \cite{Wetzstein2011, Wetzstein2012} or from a focal stack \cite{Takahashi2018}. The main difference however, is that, because of the physics of the light attenuating LCD layers, the sub-aperture images of the Light Field are reconstructed as a product of the layers' pixels instead of a sum. Hardware limitations also impose constraints on the layer representation. For instance, the number of layers is generally small (e.g. 3 to 5), which is often insufficient to accurately represent the whole Light Field. Furthermore, the layers must only have positive pixel values in order to be displayed on the LCD panels. This constraint is not required in our model, which allows us to efficiently construct the layers in the Fourier domain.


Similarly to the FDL method proposed in this paper, Alonso et al. \cite{Alonso2016} construct layers by an optimization in the Fourier domain. However, their method is limited to a focal stack input.
In this configuration, the problem is well conditioned because the input images already contain dense angular information and each constructed layer is associated to the focusing depth of one of the focal stack images. Hence, no regularization scheme was considered for this application. The method we propose is more generic as it can also construct the layers from sub-aperture images. Therefore, specific regularization strategies are studied, which allows us to address a much larger range of applications including calibration, view interpolation, denoising, etc.

Finally, the proposed FDL representation directly relates to the dimensionality gap Light Field prior described by Levin and Durand \cite{Levin2010}. It states that the support of the Light Field data in the 4D Fourier domain is a 3D manifold which was later characterized as a hypercone in \cite{Dansereau2013}.
By additionally considering the limited depth range of a scene, Dansereau et al. \cite{Dansereau2013} determined that the frequency-domain support of the Light Field forms a hyperfan. They define this shape as the intersection of the hypercone with a dual fan previously described in \cite{Dansereau2007}.
In this paper, we derive the FDL representation from the dimensionality gap prior assuming a discrete set of depths instead of a continuous range. For the discrete depth case, we show formally in Section \ref{ssec:Prior} that this prior is itself derived from the assumption of a non-occluded Lambertian scene. This is a limitation for any method directly enforcing the dimensionality gap prior. For instance, as observed in \cite{Levin2010, Dansereau2013}, in the reconstructed sub-aperture images, occluding objects may appear transparent near the occlusion boundaries.
However, semi-transparent objects and reflections on flat surfaces are accurately reproduced, which is particularly challenging for depth image based rendering methods. We also present in Section \ref{ssec:Denoising} a possible generalization of our layer model to allow a better representation of other non-lambertian effects and occlusions.



\section{Light Field notations}

\begin{figure}[t]
\centerline{\includegraphics[width=.5\linewidth]{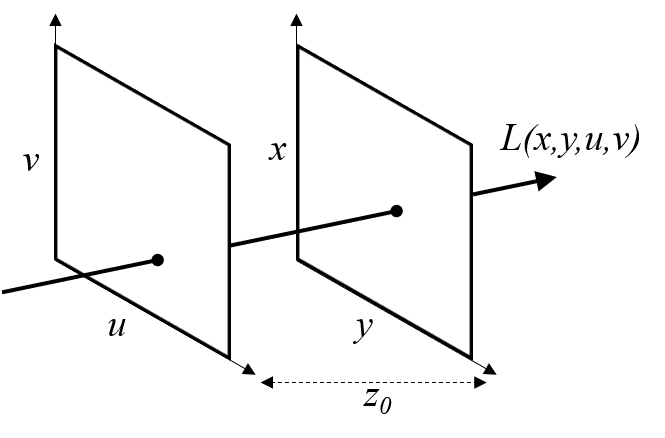}}
\caption{Two-plane parameterization. The focal plane $(x,y)$ is parallel to the camera plane $(u,v)$ and located at a distance $z_0$.}
\label{fig:2planes}
\vspace{-10pt}
\end{figure}

Let us first consider the 4D representation of Light Fields in \cite{Levoy1996} and Lumigraph in \cite{Gortler1996} parameterized with two parallel planes, as illustrated in Fig. \ref{fig:2planes}. 
The 4D representation describes the radiance along rays by a function $L(x,y,u,v)$ where the pairs $(x,y)$ and $(u,v)$ respectively represent spatial and angular coordinates.
For simplicity of notation, we consider a 2D Light Field $L(x,u)$ with one spatial dimension $x$ and one angular dimension $v$, but the generalization to a 4D Light Field $L(x,y,u,v)$ is straightforward.

In this paper, we use the notion of disparity instead of depth. Given the depth $z_0$ of the focal plane in Fig. \ref{fig:2planes}, a depth $z$ can be directly converted into a disparity $d$ with $d=\frac{z_0-z}{z}$ (i.e. objects at depth $z_0$ from the camera plane have zero disparity). Refocusing the Light Field then consists in defining a new Light Field \mbox{$L'(x,u)=L(x-us, u)$}. The refocus parameter $s$ is defined such that the regions of disparity $d=s$ in the Light Field $L$, have a disparity equal to zero in the Light Field $L'$. A refocused image, noted $B^s$, is then formed by refocusing the Light Field with parameter $s$ and integrating the light rays over the angular dimension as described in \cite{Ng2005a}:
\begin{equation}
\label{eq:refocusCentral}
B^s(x) = \int_{\mathbb{R}} L(x-us,u) \psi(u) \mathrm{d}u.
\end{equation}
where $\psi(u)$ represents the aperture of the imaging system. In the case where $B^s$ is captured by a camera with an aperture area $\mathcal{A}$ (typically a disk), then $\psi$ is defined by:
\begin{equation}
\psi(u) = \left\{
\begin{array}{ll}
        1 & \mbox{if } u\in \mathcal{A} \\
        0 & \mbox{otherwise.}
    \end{array}
\right.
\end{equation}

In this paper, we use a more general version of Eq. \eqref{eq:refocusCentral} where the image can be observed at any position $u_0$ on the camera plane with any given aperture $\psi$ :
\begin{equation}
\label{eq:refocus}
B_{u_0}^s(x) = \int_{\mathbb{R}} L(x-us,u_0+u) \psi(u) \mathrm{d}u.
\end{equation}

Note that when the aperture area is infinitely small, the function $\psi$ is equal to the dirac delta function $\delta$. In this case, the image formed by Eq. \eqref{eq:refocus} is the sub-aperture image noted $L_{u_0}$ and defined by $L_{u_0}(x)=L(x,u_0)$.

The different notations used in the article are summarized in Table \ref{tab:notations}.

\renewcommand{\arraystretch}{1.3}
\begin{table}[]
\centering
\caption{Table of notation}
\label{tab:notations}
\begin{tabular}{|c|p{.78\linewidth}|}
\hline
\textbf{Symbols}  & \textbf{Description}      \\
\hline
$L(x,u)$          & \specialcell{Light Field with spatial coordinate $x$ and angular \\ coordinate $u$.} \\
\hline
$\omega_x$, $\omega_u$  & Respectively spatial and angular frequencies.\\
\hline
$L_{u_0}$      & Sub-aperture image at position $u_0$ ($L_{u_0}(x) = L(x,u_0)$). \\
\hline
$B_{u_0}^s$       & Image with refocus parameter $s$ and position $u_0$.\\
\hline
$\psi$            & Aperture function.             \\
\hline
$u_0$             & Angular coordinate of the view to reconstruct. \\
\hline
$u_l$             & Angular coordinate of a known input view. \\
\hline
$d_k$             & Disparity value in the Light Field. \\
\hline
$\Omega_k^u$      & Region of disparity $d_k$ in an image of view position $u$. \\
\hline
$ \delta $        & Dirac delta function    \\
\hline
$\hat{f}$         & Fourier transform of a function $f$. \\
\hline
$ \matr{A}^\top$  & Transpose of the matrix $\matr{A}$. \\
\hline
$\overline{\matr{A}}$ & Complex conjugate of $\matr{A}$ (without transpose). \\
\hline
$\matr{A}^*$     & Conjugate transpose of $\matr{A}$ ($\matr{A}^* = \overline{\matr{A}^\top}$). \\
\hline
$\matr{A}_{j,k}$ & Element of $\matr{A}$ on the $j^{th}$ row and $k^{th}$ column. \\
\hline
$\matr{A}_k$     & $k^{th}$ column of matrix $\matr{A}$. \\
\hline
$\vect{x}_k$     & $k^{th}$ element of vector $\vect{x}$. \\
\hline
\end{tabular}
\end{table}
\renewcommand{\arraystretch}{1}


\section{Fourier Disparity Layer representation}
\label{sec:FDLRepresentation}

\subsection{Light Field Prior and FDL Representation}
\label{ssec:Prior}

For the derivations, we assume that the scene is lambertian, without occlusion, and can be divided into $n$ spatial regions $\Omega_k$ with constant disparity $d_k$. Formally, this can be written:
\begin{equation}
\label{eq:Prior1}
\forall k\in \ldbrack 1,n\rdbrack, \forall (x,u) \in \Omega_k\times\mathbb{R}, L(x-ud_k,u) = L(x,0).
\end{equation}

Here, the spatial regions $\Omega_k$ are defined for the central view at $u=0$. From this assumption, we prove in Appendix \ref{app:prior} that the Fourier transform of the Light Field can be decomposed as follows:
\begin{equation}
\label{eq:FourierLF}
\hat{L}(\omega_x,\omega_u) = \sum_{k} \delta(\omega_u-d_k\omega_x) \hat{L}^k(\omega_x),
\end{equation}
where $\hat{L}^k$ is defined by
\begin{equation}
\hat{L}^k(\omega_x)=\int_{\Omega_k} e^{-2i\pi x\omega_x} L(x,0) \mathrm{d}x.
\end{equation}
Each function $\hat{L}^k$ can be interpreted as the Fourier transform of the central view only considering the region $\Omega_k$ of disparity $d_k$. Hence, we call these functions Fourier Disparity Layers (FDL).
One can note from Eq. \eqref{eq:FourierLF} that the Light Field information is entirely contained in the Fourier Disparity Layers and their associated disparity values $d_k$. We will show that the layers $\hat{L}^k$ can be constructed either from sub-aperture or wide aperture images without knowing the regions $\Omega_k$. In other words, our method does not require a disparity map. However, the disparity values $d_k$ are necessary. Hence, a method for estimating these values is also presented in Section \ref{ssec:Calibration}.

Another interpretation of Eq. \eqref{eq:FourierLF} is that the Light Field is sparse in the Fourier domain and the non-zero values are located at frequencies such that $\omega_u=d_k\omega_x$ for each disparity value $d_k$ of the Light Field. This is similar to the Light Field prior referred to as dimensionality gap prior in \cite{Levin2010}, and resulting in the definition of the hyperfan filter in \cite{Dansereau2013}. However, these previous works consider a continuous disparity range, which does not allow for a practical layer representation, since it would result in an infinite number of layers.




\subsection{FDL Decomposition of Images}
\label{ssec:Derivations}


\subsubsection*{Sub-aperture images}
\label{sssec:FTView}
Here, we derive the relationship between the FDL representation of the Light Field and the Fourier Transform $\hat{L}_{u_0}$ of a sub-aperture image $L_{u_0}$. Note that in $\hat{L}_{u_0}$, the Fourier Transform only applies to the spatial dimension. Hence, $\hat{L}_{u_0}$ is obtained from the Light Field's spectrum $\hat{L}$ by applying the inverse Fourier transform in the angular dimension as follows:

\begin{equation}
\hat{L}_{u_0}(\omega_x) = \int_{-\infty}^{+\infty} e^{+2i\pi u_0\omega_u} \hat{L}(\omega_x,\omega_u)\mathrm{d}\omega_u.
\label{eq:FourierView0}
\end{equation}

Using the FDL decomposition of $\hat{L}(\omega_x,\omega_u)$ from Eq. \eqref{eq:FourierLF}, the transformed sub-aperture image can be directly derived:
\begin{flalign}
\quad &\hat{L}_{u_0}(\omega_x) &\\
&\quad = \sum_k \hat{L}^k(\omega_x)\int_{-\infty}^{+\infty} e^{+2i\pi u_0\omega_u} \delta(\omega_u-d_k\omega_x)\mathrm{d}\omega_u &\\
&\quad = \sum_k e^{+2i\pi u_0 d_k\omega_x}\hat{L}^k(\omega_x).
\label{eq:FourierView}
\end{flalign}

\subsubsection*{General image model}
\label{sssec:FTBs}
Now, considering the general model in Eq. \eqref{eq:refocus} the Fourier Transform of an image $B_{u_0}^s$ is:
\begin{flalign}
\begin{split}
\ & \hat{B}_{u_0}^s(\omega_x) \\
&\,  = \int_{-\infty}^{+\infty} e^{-2i\pi x\omega_x}\left[ \int_{-\infty}^{+\infty}L(x-us,u_0+u) \psi(u) \mathrm{d}u \right]\mathrm{d}x
\end{split} &\\
&\,  = \int_{-\infty}^{+\infty} e^{-2i\pi (x+us)\omega_x} \psi(u) \int_{-\infty}^{+\infty} L(x,u_0+u)\mathrm{d}x  \mathrm{d}u &\\
&\,  = \int_{-\infty}^{+\infty} e^{-2i\pi us\omega_x} \psi(u) \hat{L}_{u_0+u}(\omega_x)\mathrm{d}u.
\label{eq:BsIntermediate}
\end{flalign}

Using the FDL decomposition of $\hat{L}_{u_0+u}$ from Eq. \eqref{eq:FourierView}, we obtain:

\begin{flalign}
\begin{split}
\ & \hat{B}_{u_0}^s(\omega_x) \\
&\, = \int_{-\infty}^{+\infty} e^{-2i\pi us\omega_x} \psi(u) \sum_k e^{+2i\pi (u_0+u) d_k\omega_x}\hat{L}^k(\omega_x)\mathrm{d}u 
\end{split} &\\
&\, = \sum_k e^{+2i\pi u_0 d_k\omega_x} \hat{L}^k(\omega_x) \int_{-\infty}^{+\infty} e^{2i\pi u(d_k-s)\omega_x} \psi(u)\mathrm{d}u &\\
&\, = \sum_k e^{+2i\pi u_0 d_k\omega_x} \hat{\psi}(\omega_x(s-d_k))\cdot \hat{L}^k(\omega_x).
\label{eq:BsFourier}
\end{flalign}
One can easily verify that in the particular case of an infinitely small aperture such that $\psi=\delta$, we have $\hat{\psi}=1$, thus Eq. \eqref{eq:BsFourier} becomes equivalent to the sub-aperture case in Eq. \eqref{eq:FourierView}.

\section{Construction of the Layers}
\label{sec:LayerConstruction}

The layer construction is performed from a set of $m$ images noted $\{B_j\}_{j\in \ldbrack 1,m\rdbrack}$ that follow the model of Eq. \eqref{eq:refocus} (e.g. images captured by cameras on the same plane and oriented perpendicularly to this plane). Each image $B_j$ is associated with its angular coordinate $u_j$, aperture function $\psi_j$, and refocus parameter $s_j$.
Thanks to the genericity of this model, different forms of input data may be used. For example, in the case of a set of sub-aperture images, the aperture function $\psi_j$ of each image is equal to the dirac delta function $\delta$. Note that, in this case, the refocus parameter $s_j$ has no influence in the equations (e.g. see Eq. \eqref{eq:FourierView}) and is not required.
Another notable example is that of a focal stack input (i.e. images with the same angular coordinates $u_j$ and aperture functions $\psi_j$, but with different refocus parameters $s_j$).

The layer construction problem is formulated in the general case in the next sub-section, assuming all the image parameters are known. Then, in sub-section \ref{ssec:Calibration}, we show how to determine the positions $u_j$ of the views and the layer's disparity values $d_k$ when the input data is a set of sub-aperture images.

\subsection{Problem Formulation}
\label{ssec:Formulation}

The results of Eqs. \eqref{eq:FourierView} and \eqref{eq:BsFourier} show that for a fixed spatial frequency $\omega_x$ in the Fourier domain, the images $B_j$ can be simply decomposed as linear combinations of the Fourier Disparity Layers. Therefore, the FDL representation of the Light Field can be learned by linear regression for every coefficient in the discrete Fourier domain.

We first compute the discrete Fourier Transform $\hat{B}_j$ of each image $B_j$ and we construct, for each frequency component $\omega_x$, a vector $\vect{b}$ such that $\vect{b}_j = \hat{B}_j(\omega_x)$.
Given the disparity value $d_k$  of each layer ($k\in \ldbrack 1,n\rdbrack$), we also construct the matrix $\matr{A} \in \mathbb{C}^{m\times n}$ with:

\begin{equation}
\label{eq:AMatGeneral}
\matr{A}_{jk}=e^{+2i\pi u_j d_k\omega_x} \hat{\psi_j}(\omega_x(s_j-d_k)).
\end{equation}

By defining the vector $\vect{x}$ with $\vect{x}_k = \hat{L}^k(\omega_x)$, the FDL decomposition in Eq. \eqref{eq:BsFourier} is reformulated as $\matr{A}\vect{x}=\vect{b}$.
A simple layer construction method then consists in solving an ordinary linear least squares problem independently for each frequency component $\omega_x$.
In practice, however, this may be an ill-posed problem. It is typically the case when the number of available images is lower than the number of layers required to represent the scene (i.e. $m<n$). More generally, depending on the input configuration, and for some frequency components, the matrix $\matr{A}$ may be ill-conditioned. This results in over-fitting and extreme noise amplification when the layer model is used to render new images of the scene (e.g. view interpolation or extrapolation). In order to avoid this situation, we include a Tikhonov regularization term in the problem formulation:
\begin{equation}
\label{eq:LayerConstructProb}
\vect{x} = \argmin_\vect{x}{\norm{\matr{A}\vect{x}-\vect{b}}_2^2 + \lambda\norm{\matr{\Gamma}\vect{x}}_2^2},
\end{equation}
where $\matr{\Gamma}$ is the Tikhonov matrix, and $\lambda$ is a parameter controlling the amount of regularization.
The problem \eqref{eq:LayerConstructProb}, has a well-known closed form solution:
\begin{equation}
\label{eq:LayerConstructionClosedForm}
    \vect{x} = (\matr{A}^* \matr{A} + \lambda \matr{\Gamma^* \Gamma})^{-1} \matr{A}^*\vect{b},
\end{equation}
where $^*$ is the Hermitian transpose operator (i.e. complex conjugate of the transposed matrix).

In our implementation of the layer construction, the Tikhonov matrix is defined according to the $2^{nd}$ order view regularization scheme presented in subsection \ref{ssec:RegularizationView} (see Eq. \eqref{eq:2ndOrderView}). It encourages smooth variations between close viewpoints generated from the FDL model, which is intuitively a desirable property for a Light Field.


 
\subsection{FDL Calibration}
\label{ssec:Calibration}

We additionally propose a calibration method that jointly estimates the layers' disparity values $d_k$ and the angular coordinates $u_j$ of the input images $B_j$. For simplicity, the calibration is restricted to the case of sub-aperture images with infinitely small aperture such that $\psi_j=\delta$. In this case, Eq. \eqref{eq:AMatGeneral} has a simpler expression \mbox{$\matr{A}_{jk}=e^{+2i\pi u_j d_k\omega_x}$}. Note that in the general configuration where the aperture is unknown, both the aperture functions $\psi_j$ and the refocus parameters $s_j$ would also need to be estimated. However, that generalization is out of the scope of this paper and is left for future work.

In what follows we express $\matr{A}$ as a matrix function
\mbox{$\matr{A} \colon \mathbb{R}^{m\times n} \to \mathbb{C}^{m\times n}$} such that, for a matrix $\matr{M}\in\mathbb{R}^{m\times n}$, \mbox{$\matr{A}(\matr{M})_{j,k} = e^{+2i\pi \matr{M}_{j,k}}$}.
The calibration problem can then be stated in a similar way as the layer construction problem \eqref{eq:LayerConstructProb} by treating the calibration parameters as unknowns in the minimization. However, these parameters do not depend on the frequency. Hence, the function to minimize is expressed as a sum over the $Q$ frequency components $\omega_x^q$ ($Q$ is equal to the number of pixels in each input image):
\begin{equation}
\label{eq:CalibrationProb}
\min_\vect{x,u,d}{\sum_{q=1}^Q\left(\norm{\matr{A}(\omega_x^q\vect{u d^\top})\vect{x}^q-\vect{b}^q}_2^2 + \lambda\norm{\matr{\Gamma}\vect{x}^q}_2^2\right)},
\end{equation}
where the input view positions $u_j$ and the disparity values $d_k$ are arranged in the column vectors $\vect{u}$ and $\vect{d}$ respectively \mbox{($\vect{u d^\top}\in\mathbb{R}^{m\times n}$)}. The vectors $\vect{x}^q$ and $\vect{b}^q$ contain the Fourier coefficients of, respectively, the disparity layers and the input images at the frequency $\omega_x^q$ (i.e. \mbox{$\vect{x}_k^q = \hat{L}^k(\omega_x^q)$ and} \mbox{$\vect{b}_j^q=\hat{B}_j(\omega_x^q)$)}.
Unlike the layer construction problem, the matrix $\matr{\Gamma}$ is defined according to the $2^{nd}$ order layer regularization approach detailed in subsection \ref{ssec:RegularizationLayer} (see Eq. \eqref{eq:2ndOrderLayer}).

In order to solve this problem, we perform a gradient descent along the vectors of parameters $\vect{u}$ and $\vect{d}$. At each iteration, the current estimate of $\vect{u}$ and $\vect{d}$ is first used to update the layers values in each vector $\vect{x}^q$ using Eq. \eqref{eq:LayerConstructionClosedForm}. The layers values are then used to compute the gradients $\nabla\vect{u}$ and $\nabla\vect{d}$ of the objective function in Eq. \eqref{eq:CalibrationProb} along $\vect{u}$ and $\vect{d}$ respectively.

By differentiation with respect to each element of $\vect{u}$ and $\vect{d}$, one can show that the gradients are expressed as:
\begin{align}
\nabla\vect{u}_j &= \sum_k \vect{d}_k \nabla \matr{P}_{j,k} \\
\label{eq:gradD}
\nabla\vect{d}_k &= \sum_j \vect{u}_j \nabla \matr{P}_{j,k},
\end{align}
with $\matr{P}=\vect{ud^\top}$, and each column $k$ of the corresponding gradient matrix $\nabla\matr{P}$ is:
\begin{equation}
\label{eq:gradP}
\nabla\matr{P}_k = 4\pi\sum_q \operatorname{Im}\!\Big(\omega_x^q \overline{\vect{x}_k^q}\cdot \overline{\matr{A}(\omega_x^q\matr{P})}_k \circ (\matr{A}(\omega_x^q\matr{P})\vect{x}^q-\vect{b}^q) \Big),
\end{equation}
where $\circ$ is the Hadamard product (i.e. element-wise multiplication), and $\operatorname{Im}$ is the imaginary part. Note that the matrix $\matr{A}(\omega_x^q\matr{P})$ was previously constructed to determine $\vect{x}^q$, and can thus be re-used in Eq. \eqref{eq:gradP} to efficiently compute the gradients.

The computations are further accelerated by performing a stochastic gradient descent, where only a small subset of the $Q$ frequency components $\omega_x^q$ is selected randomly at each iteration for the gradients computation. In our implementation, subsets of 4096 frequency components were selected. Therefore, the computational cost per iteration does not depend on the image resolution.

Finally, given the gradients, the updated vectors $\vect{u'}$ and $\vect{d'}$ are then computed as
\begin{equation}
\vect{u'} = \vect{u}-\alpha\frac{\nabla\vect{u}}{\epsilon + \norm{\nabla\vect{u}}_2^2}
,\ and\quad
\vect{d'} = \vect{d}-\alpha\frac{\nabla\vect{d}}{\epsilon + \norm{\nabla\vect{d}}_2^2},
\end{equation}
where $\epsilon$ is a small value encouraging the stability of the algorithm when the gradients become small (i.e. when $\vect{u}$ and $\vect{d}$ are close to an optimum). In our experiments a fixed value $\alpha=0.2$ was used.


\subsection{Regularization Schemes}
\label{ssec:Regularization}
\begin{figure}
\centering
\begin{minipage}[h]{.397\linewidth}
\centerline{\includegraphics[width=\linewidth]{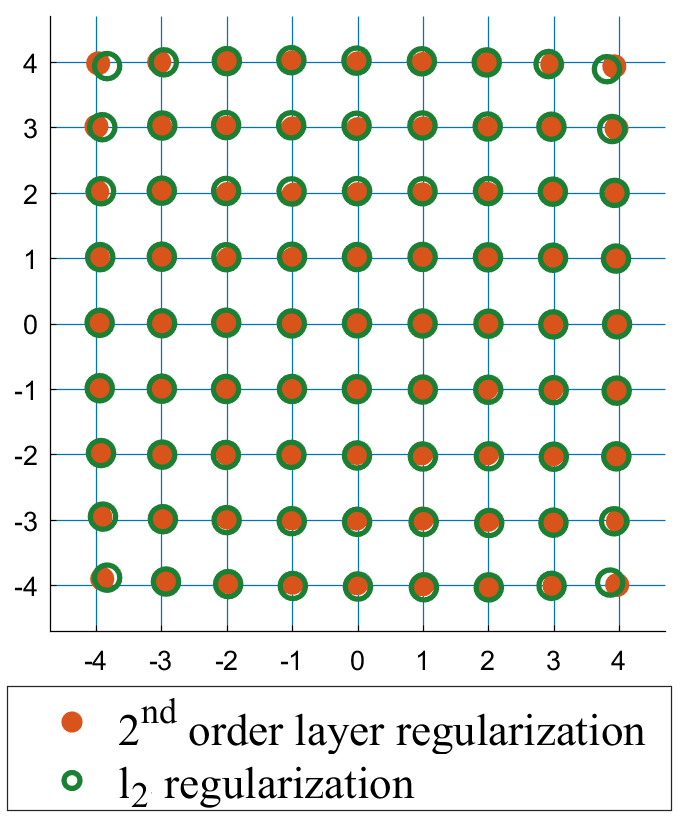}}
\vspace{-3pt}
\centerline{\small{(a)}}
\vspace{-5pt}
\end{minipage}
\begin{minipage}[h]{.583\linewidth}
\centerline{\includegraphics[width=\linewidth]{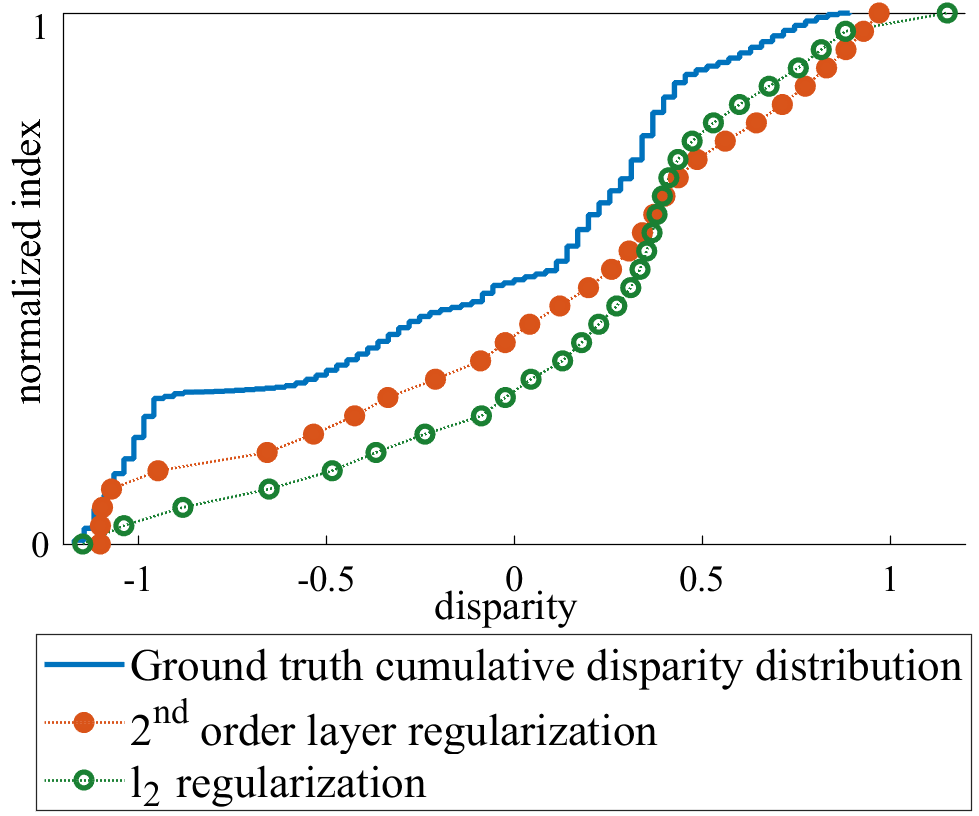}}
\vspace{-3pt}
\centerline{\small{(b)}}
\vspace{-3pt}
\end{minipage}
\caption{Example of calibration results for the synthetic Light Field `papillon' \cite{wanner2013datasets} with 30 layers: (a) Estimated angular coordinates of the views (the ground truth is the 9x9 regular grid). (b) Sorted disparity values. The vertical axis represents the normalized index of each of the sorted values. The blue curve was obtained by sorting all the pixels of the ground truth disparity map in order to visualize the cumulative disparity distribution.}
\label{fig:Calib}
\vspace{-6pt}
\end{figure}

In order to keep the layer construction and calibration problems in Eqs. \eqref{eq:LayerConstructProb} and \eqref{eq:CalibrationProb} easy to solve, we have used a Tikhonov regularization. The definition of the Tikhonov matrix $\matr{\Gamma}$ depends on the intended objective.
In the most basic form, classical $l_2$ regularization is obtained by simply taking $\matr{\Gamma}$ equal to the identity matrix $\matr{I}_n$. This prevents the values in $\vect{x}$ (i.e. Fourier coefficients of the disparity layers) from taking too high values, which reduces the noise, but also results in a loss of details. Furthermore, for the calibration problem, using the $l_2$ regularization may not accurately estimate the disparity distribution of the Light Field as illustrated by the calibration results of Fig. \ref{fig:Calib}. The sorted disparity values shown in Fig. \ref{fig:Calib}(b) can be interpreted as an estimation of the cumulative disparity distribution.

In this section, we present two schemes referred to as $2^{nd}$ order view regularization and $2^{nd}$ order layer regularization which are better suited to the layer construction and the calibration respectively.

\subsubsection{$2^{nd}$ Order View Regularization}
\label{ssec:RegularizationView}
For the layer construction, we want to encourage smooth variations between close viewpoints generated from the layers. For that purpose, at each frequency, we penalize the second derivative of sub-aperture images generated from the model with respect to the angular coordinate.
From the expression of sub-aperture images in Eq. \eqref{eq:FourierView}, the second derivative at a coordinate $u_0$ is given by:
\begin{equation}
\frac{\partial^2 \hat{L}_{u}(\omega_x)}{\partial u^2}\Bigr|_{u=u_0}  = -4\pi^2\sum_k{w_x}^2{d_k}^2 e^{+2i\pi u_0 d_k\omega_x}\hat{L}^k(\omega_x).
\label{eq:2ndDer}
\end{equation}

Let us now consider a set \mbox{$\mathcal{R}$} of angular coordinates to regularize. From Eq. \eqref{eq:2ndDer}, the $2^{nd}$ order view regularization then consists in constructing a matrix $\matr{\Gamma}$ where each row $l$ is associated to an angular coordinate $u_l\in\mathcal{R}$ and is defined by \mbox{$\matr{\Gamma}_{l,k} = {w_x}^2{d_k}^2 e^{+2i\pi u_l d_k\omega_x}$}. For simplicity, we ignore the constant factor $-4\pi^2$ in the definition of $\matr{\Gamma}$ since it can be accounted for in the regularization parameter $\lambda$. Intuitively, in order to apply the regularization at every coordinate in the camera plane, one would need to define a matrix $\matr{\Gamma}$ with an infinite number of rows (i.e. infinite set $\mathcal{R}$). Although this might seem impractical, we show in what follows that it can be done by observing that the solution to the regularized problem in Eq. \eqref{eq:LayerConstructionClosedForm} only requires the knowledge of $\matr{\Gamma}^*\matr{\Gamma}$ of finite size $n\times n$. 

First, let us take a continuous interval $\mathcal{R}=[-r/2,r/2]$ of size $r$, instead of a discrete set. Then, each column $\matr{\Gamma}_k$ must be interpreted instead as a function $\Gamma_k : u \mapsto {w_x}^2{d_k}^2 e^{+2i\pi u d_k\omega_x}$. Thus, each element of $\matr{\Gamma}^*\matr{\Gamma}$ is given by the following inner product:
\begin{align}
\label{eq:GTG}
[\matr{\Gamma}^*\matr{\Gamma}]_{k_1,k_2} &= \langle\Gamma_{k_1}, \Gamma_{k_2}\rangle = \int_{-r/2}^{r/2} \Gamma_{k_1}(u)\cdot\widebar{\Gamma_{k_2}}(u) \mathrm{d}u, &\\
&= {\omega_x}^4{d_{k_1}}^2{d_{k_2}}^2 \int_{-r/2}^{r/2} e^{2 i\pi u({d_{k_1}}-{d_{k_2}})\omega_x}\mathrm{d}u, &\\
&= {\omega_x}^4{d_{k_1}}^2{d_{k_2}}^2 r\, \mathrm{sinc}(r({d_{k_1}}-{d_{k_2}})\omega_x).
\end{align}
For convenience, we replace $\matr{\Gamma}^*\matr{\Gamma}$ in Eq. \eqref{eq:LayerConstructionClosedForm} by the scaled version $\widetilde{\matr{\Gamma}^*\matr{\Gamma}} = \frac{\matr{\Gamma}^*\matr{\Gamma}}{r}$, so that the amount of regularization remains independent of the size of the integration domain in Eq. \eqref{eq:GTG} (i.e. area of the portion of the camera plane covered by the regularization).
In our implementation of the $2^{nd}$ order view regularization, we consider the full camera plane (i.e. $\mathcal{R}=(-\infty,\infty)$) by making $r$ tend towards the infinity. In this case, $\widetilde{\matr{\Gamma}^*\matr{\Gamma}}$ simply becomes a diagonal matrix such that:
\begin{equation}
\label{eq:2ndOrderView}
[\widetilde{\matr{\Gamma}^*\matr{\Gamma}}]_{k,k} = {\omega_x}^4{d_k}^4.
\end{equation}

Note that for $\omega_x=0$, Eq. \eqref{eq:2ndOrderView} gives $\widetilde{\matr{\Gamma}^*\matr{\Gamma}}=0$, which does not produce any regularization. Therefore, in practice, a small value $\epsilon$ is added to the diagonal elements of $\widetilde{\matr{\Gamma}^*\matr{\Gamma}}$. This is equivalent to adding a $l_2$ regularization term $\epsilon\cdot\norm{\vect{x}}_2^2$ to the problem \eqref{eq:LayerConstructProb}.

\subsubsection{$2^{nd}$ Order Layer Regularization}
\label{ssec:RegularizationLayer}
For the calibration step, however, it is preferable to use a regularization that does not depend on the parameters $d_k$. Otherwise, the expression of the gradients $\nabla\vect{d}$ would need to take the regularization term into account, which can result in more complex gradient computations and a slower convergence of the algorithm. Instead, we encourage smooth variations between successive layers by defining $\matr{\Gamma}$ as a discrete approximation of the second-order differential operator as follows:


\begin{equation}
\label{eq:2ndOrderLayer}
\matr{\Gamma}=
\begin{bmatrix}
-2  &    1   &        &        &     \\
 1  &   -2   &    1   &        &     \\
    & \ddots & \ddots & \ddots &     \\
    &        &    1   &   -2   &  1  \\
    &        &        &    1   & -2  \\
\end{bmatrix}.
\end{equation}
The $2^{nd}$ order layer regularization thus penalizes large differences between neighboring layers, which results in a more uniform distribution of the disparity values. As shown in Fig. \ref{fig:Calib}(b), the calibration using the $l_2$ regularization (i.e. $\matr{\Gamma}=\matr{I}_n$) tends to find too many disparity values close to the dominant disparity (between 0 and 0.5 in the figure) and may underestimate other parts of the light field (e.g. with disparities close to -1). The $2^{nd}$ order layer regularization attenuates that effect by encouraging a more uniform distribution.
Note, however, that for the layer construction, smoothness along the layers is not desirable and produces artifacts along the edges in the images rendered from the FDL model.


\section{Light Field Rendering}
\label{ssec:Rendering}

\subsection{Implementation}

\begin{figure}
\centerline{\includegraphics[width=\linewidth]{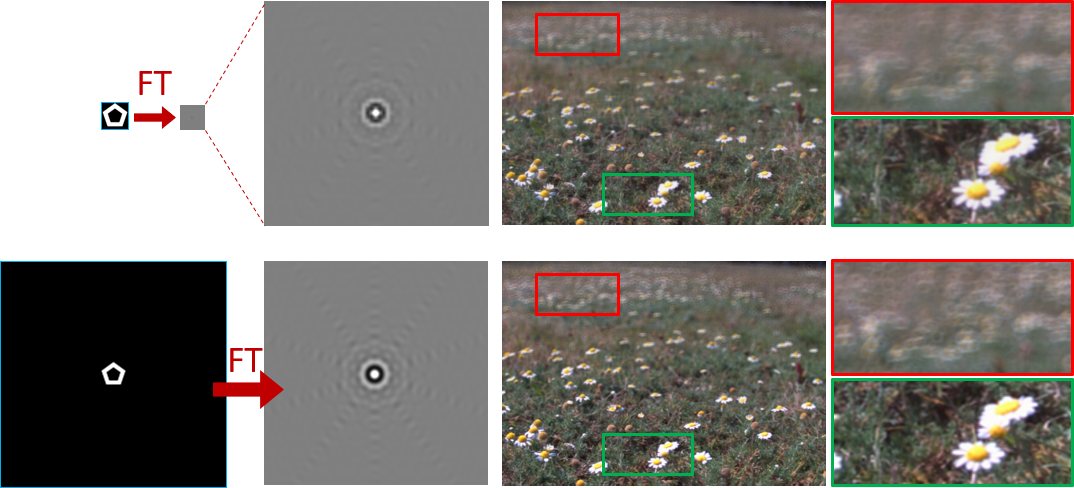}}
\vspace{-6pt}
\caption{Top row: no padding. Bottom row: with zero-padding. From left to right: Aperture image; Real part of the Fourier transform of the aperture image (up-sampled with linear interpolation for the top row for the comparison); Rendered result with magnified details. Zero-padding removes the aliasing in the Fourier transform of the aperture, which better preserves contrasts in the final rendered image (best viewed zoomed in).}
\label{fig:apPad}
\vspace{-6pt}
\end{figure}

Knowing the layers and their disparities, any view with arbitrary aperture and focus can be rendered in the Fourier domain by applying the FDL decomposition equation \eqref{eq:BsFourier} and by computing the inverse Fourier transform. The interpretation in the spatial domain is that each layer is shifted (i.e. multiplied by $e^{+2i\pi u_0 d_k \omega_x}$ in the Fourier domain) and filtered (i.e. multiplied by $\hat{\psi}(\omega_x(s-d_k))$ in the Fourier domain).

However, except for specific aperture shapes (e.g. square or dirac), the Fourier transform $\hat\psi$ of the aperture function has no analytical expression. Hence, in our implementation, an approximation is obtained by drawing the aperture shape as an image, and by taking its discrete Fourier transform. A linear interpolation is used to determine $\hat{\psi}(\omega_x(s-d_k))$ from the discrete frequency samples. In order to increase the accuracy of this approximation, the aperture image is zero-padded before computing its Fourier transform, which increases the resolution in the spectral domain. The effect of the padding is illustrated in Fig. \ref{fig:apPad} showing that the resulting increased  spectral domain resolution allows a better preservation of the contrasts in the rendered images.
Note also that the spatial resolution of the aperture image can be taken arbitrarily large without affecting the aperture size in the rendered image. In practice, we control the aperture size by replacing $\hat{\psi}(\omega_x(s-d_k))$ in Eq. \eqref{eq:BsFourier} by $\hat{\psi}(f\omega_x(s-d_k))$ with a scaling factor $f$. For example, taking $f=0$ simulates a sub-aperture image. On the other hand, taking a large factor $f$ produces a shallow depth of field effect without affecting the complexity of the method.

\subsection{Visual Results and Discussion}

\begin{figure}
\centering
\begin{minipage}[h]{1\linewidth}
\centerline{\includegraphics[width=\linewidth]{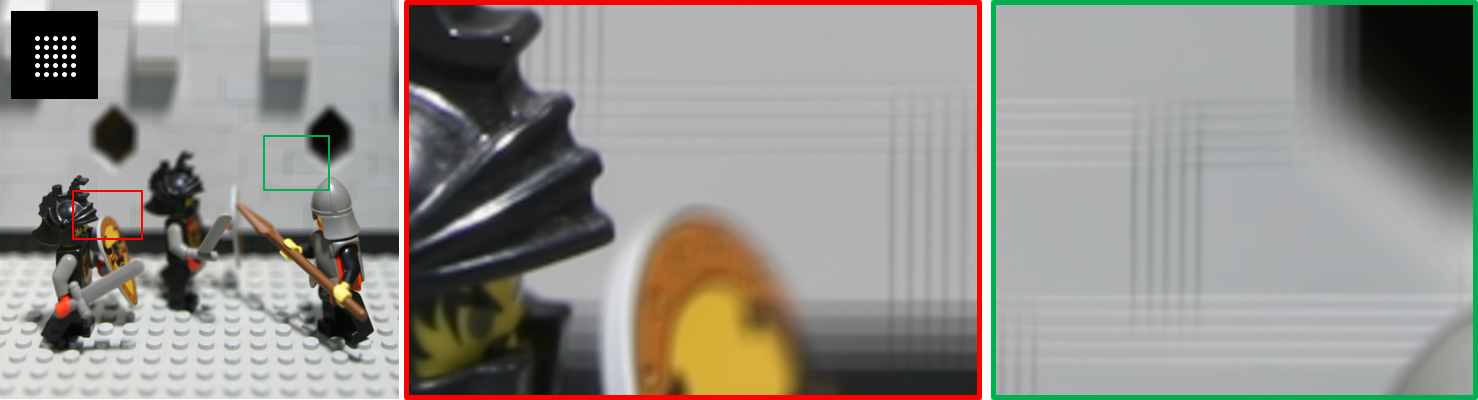}}
\vspace{-3pt}
\centerline{\small{(a)}}
\vspace{3pt}
\end{minipage}
\begin{minipage}[h]{1\linewidth}
\centerline{\includegraphics[width=\linewidth]{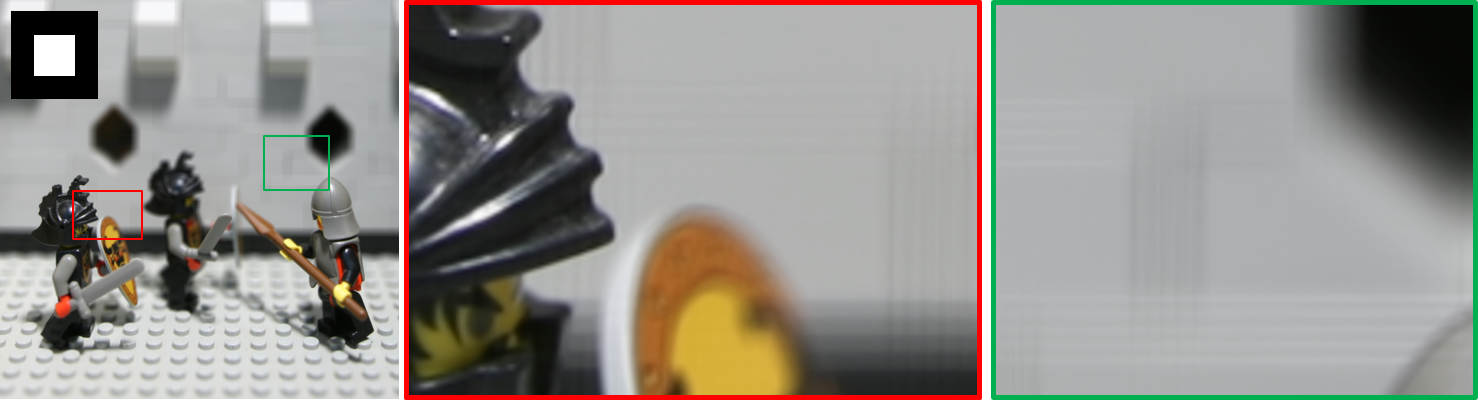}}
\vspace{-3pt}
\centerline{\small{(b)}}
\vspace{3pt}
\end{minipage}
\begin{minipage}[h]{1\linewidth}
\centerline{\includegraphics[width=\linewidth]{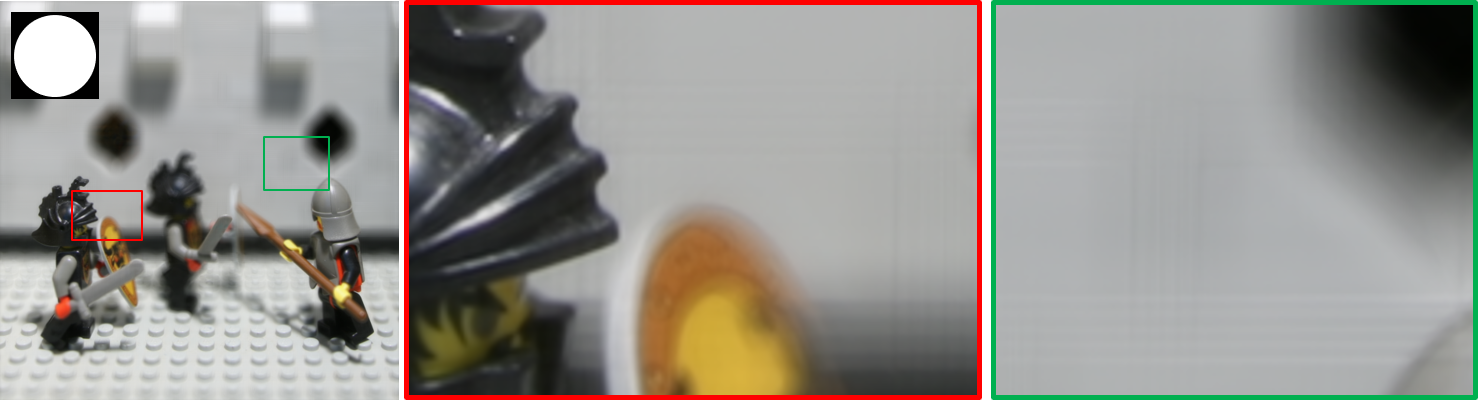}}
\vspace{-3pt}
\centerline{\small{(c)}}
\end{minipage}
\vspace{-10pt}
\caption{Refocused image using 5x5 views of the Light Field `Lego Knights' from the Stanford dataset \cite{StanfordDataset}: (a) Shift and Sum method \cite{Ng2005a}, (b) Our method using a square aperture covering the angular coordinates of the input 5x5 views. (c) Our method using a larger circular aperture. Angular aliasing is reduced in our method by considering all the angular coordinates within the aperture.}
\label{fig:OursVsShiftAdd}
\vspace{-6pt}
\end{figure}

In Fig. \ref{fig:OursVsShiftAdd}, we show  a comparison of our approach with the conventional Shift and Sum refocusing method \cite{Ng2005a}. Thanks to the regularization used in the layer construction (see Section \ref{ssec:Regularization}), the light field is extended angularly, which allows rendering images with reduced angular aliasing as in Fig. \ref{fig:OursVsShiftAdd}(b). As shown in Fig. \ref{fig:OursVsShiftAdd}(c), the aperture can also be taken larger than the baseline of the input Light Field from which the FDL model is constructed. The possibility to change the shape of the aperture also gives better control over the bokeh (i.e. the quality of the out-of-focus regions).

\begin{figure}
\centering
\begin{minipage}[h]{.88\linewidth}
\centerline{\includegraphics[width=\linewidth]{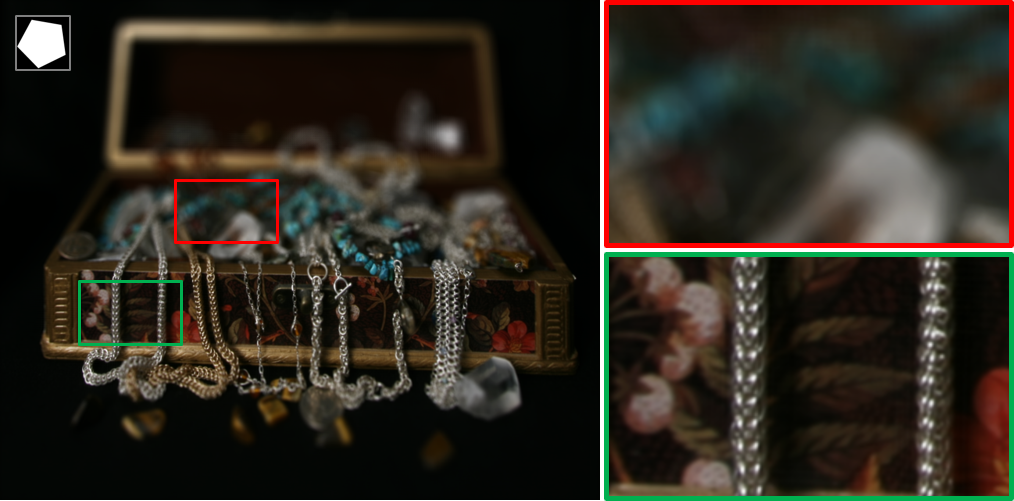}}
\vspace{-3pt}
\centerline{\small{(a)}}
\vspace{3pt}
\end{minipage}
\begin{minipage}[h]{.88\linewidth}
\centerline{\includegraphics[width=\linewidth]{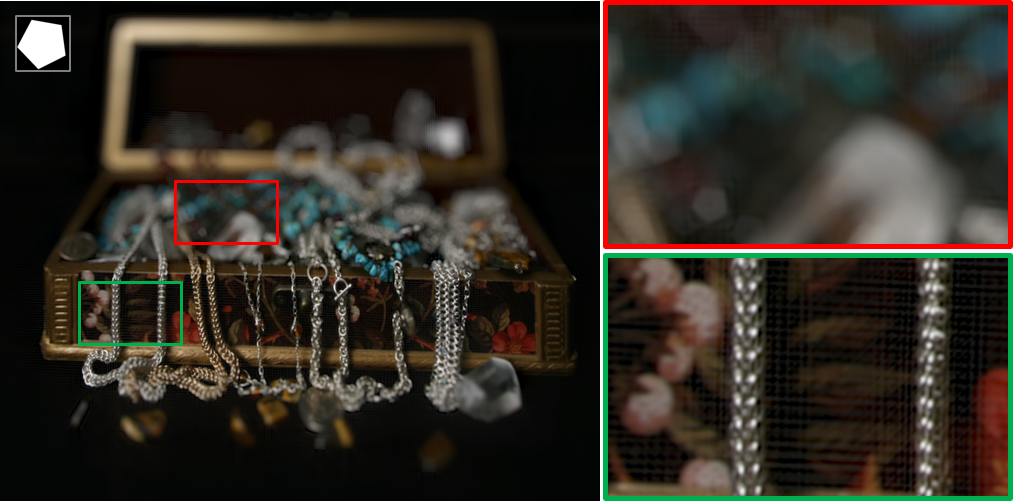}}
\vspace{-3pt}
\centerline{\small{(b)}}
\end{minipage}
\vspace{-5pt}
\caption{Rendered result when the processing is performed in (a) gamma corrected space, (b) linear space. Linear space processing results in brighter out-of-focus regions where the bokeh shape (i.e. aperture shape shown in the top left corner) appears more clearly. However, it also increases the artifacts (see details in the green rectangles).}
\label{fig:GammaVsLinear}
\vspace{-6pt}
\end{figure}

Another important and often overlooked factor influencing the bokeh is the color space of the input image data. The refocusing Eqs. \eqref{eq:refocusCentral} and \eqref{eq:refocus} assume that the Light Field rays $L(x,y,u,v)$ are proportional to luminance data. However, images are traditionally represented in a non-linear color space (e.g. gamma corrected) in order to account for the non-linearity of the human visual system as well as typical displays. This non-linearity affects the appearance of the refocused images. More realistic bokeh can be achieved by applying inverse gamma correction to the input images before the layer construction, and by applying gamma correction back after the rendering step, as shown in Fig. \ref{fig:GammaVsLinear}. One limitation of this approach, however, is that computing the layers in a linear color space also produces more visible artifacts. This is due to the fact that solving the layer construction problem \eqref{eq:LayerConstructProb} in a linear color space does not take the non-linearity of human perception of luminance into account.

A demonstration of our rendering application is available in the supplementary materials.
It shows that our approach can be used for controlling simultaneously the viewpoint, the aperture shape and size, and the focus in real-time.

\section{Complexity}
In this section, we analyse the complexity of the different processing steps proposed in our approach.

First, regarding the spatial resolution, our implementation takes advantage of the symmetry property of the Fourier transform for real signals.
Given a real valued function $g$, its Fourier transform $\hat{g}$ is such that \mbox{$\hat{g}(-\omega) = \overline{\hat{g}(\omega)}$}. Hence, for the layer construction and the rendering algorithms, only half of the discrete spectrum must be computed. The remaining frequencies are directly obtained by copying and taking the complex conjugate of the first half.
This symmetry property is also used in the calibration step by selecting random frequencies only in one half of the spectrum for the stochastic gradient descent.
Regarding memory requirements, using only half of the frequencies in the Fourier representation compensates the fact that complex numbers require twice as much memory as real numbers.

\begin{figure}
\centering
\begin{minipage}{1\linewidth}
\centerline{\includegraphics[width=\linewidth]{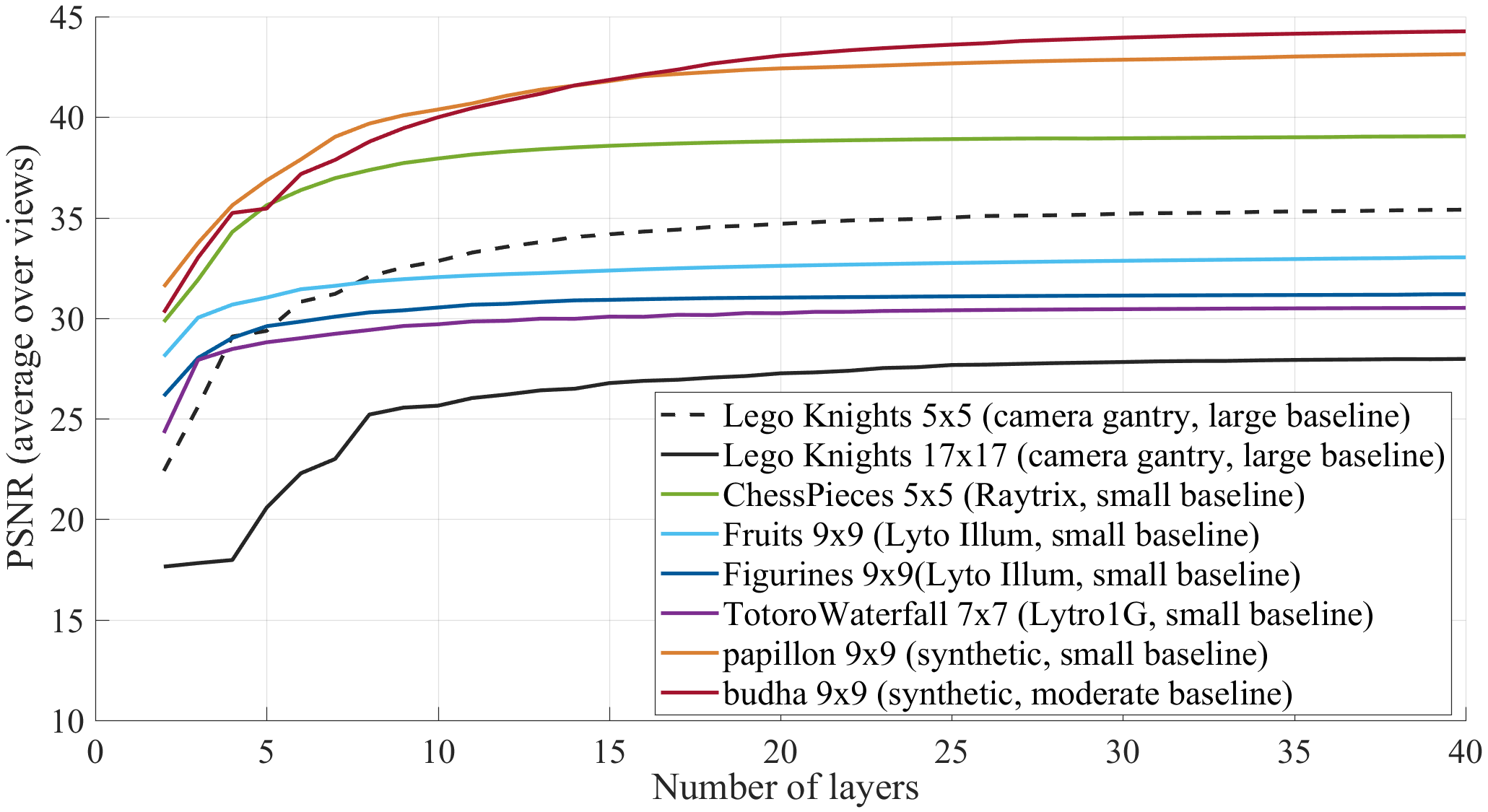}}
\end{minipage}
\vspace{-5pt}
\caption{Reconstruction PSNR (average over views) for different Light Fields with respect to the number of layers used in the FDL model. For the Light Field `Lego Knights', two different results are shown using either the full 17x17 views or only the 5x5 central views.}
\label{fig:PSNRvsLayers}
\vspace{-6pt}
\end{figure}

In addition to the spatial resolution, another important factor to consider for the complexity is the number of layers used.
In order to analyse the number of layers needed for our model, we have performed an experiment where Light Field sub-aperture images are used for FDL calibration and construction with varying numbers of layers. The input images are then reconstructed via FDL rendering. Several Light Fields from various datasets were tested including natural Light Fields captured with a first generation Lytro camera (`Totoro Waterfall' \cite{HLRA-JSTSP}), a Lytro Illum camera (`Fruits' and `Figurines' \cite{LFInpaintTIP}), a Raytrix R8 camera (first frame of the Light Field `ChessPieces' \cite{RaytrixINRIADataset}), a traditional camera moving on a gantry (`Lego Knights' \cite{StanfordDataset}), and synthetic Light Fields (`papillon' and `budha' \cite{wanner2013datasets}).
Fig. \ref{fig:PSNRvsLayers} presents the average Peak Signal to Noise Ratio (PSNR) of all the reconstructed sub-aperture images with respect to the number of layers.
The saturation of the PSNR curves show that 15 to 30 layers are generally sufficient, and using more layers does not significantly change the results. Light Fields with a large baseline, and thus large differences between extreme viewpoints (e.g. Lego 5x5, Lego 17x17), typically require more layers than smaller baseline Light Fields. However a very large range of viewpoints also implies large occlusions, which results in a significant loss when reconstructing the sub-aperture images, as observed with Lego 5x5 and Lego 17x17. This makes the analysis difficult for very large baseline Light Fields which may still benefit from more than 30 layers in the ideal non-occluded case. Note that for the small baseline Light Fields captured with Lytro cameras, the relatively low PSNR is due to inaccuracies in the input data (e.g. noise, color differences between views) which are reduced in our reconstruction. More details on the noise reduction as well as an extension for better occlusion handling are presented in section \ref{ssec:Denoising}.

The computing times for the different processing steps are presented in Table \ref{tab:ComputingTimes} for all the Light Fields in Fig. \ref{fig:PSNRvsLayers}. The table also provides the number of iterations needed for the calibration since it has a large impact on the computing time. For this experiment, 30 layers were used although less layers would be sufficient for some of the Light Fields (e.g. see `ChessPieces', `Figurines', `Totoro Waterfall' in Fig. \ref{fig:PSNRvsLayers}). The processing times were measured using an Intel Core i7-7700 CPU and a Nvidia GeForce GTX 1080 GPU. All the proposed steps were implemented in Matlab using the parallel computing toolbox in order to process the different frequencies in parallel on the GPU.

Although the processing time for the calibration and layer construction steps is affected by the number of input views (i.e. angular resolution), the rendering time only depends on the spatial resolution and the number of layers. For example, the same rendering times is obtained for `Lego Knights' using either 5x5 or 17x17 input views.  This is particularly advantageous when rendering images with a large aperture size that are traditionally obtained by averaging a large number of sub-aperture images with the Shift and Sum algorithm \cite{Ng2005a}. Note that similarly to our method, the Fourier Slice refocusing algorithm \cite{Ng2005b} first transforms the Light Field in an intermediate representation in order to perform fast refocusing. In that method the refocusing complexity does not depend on the number of input views. However, since the intermediate representation is the 4D Fourier transform of the Light Field, the memory requirement remains proportional to the number of views. Furthermore, it does not allow changing the aperture or the viewpoint.

\begin{table}[]
\centering
\caption{Computing times for our Matlab implementation using an Intel Core i7-7700 CPU and an Nvidia GeForce GTX 1080 GPU. For every Light Field 30 layers were used.}
\vspace{-5pt}
\label{tab:ComputingTimes}
\renewcommand{\tabcolsep}{1.5mm}
\begin{tabular}{|l|c|c|c|}
\hline
\textbf{Input Light Field} (resolution) & \specialcell{Calibration \\ time (\#iter)}   & \specialcell{Layer \\ construction}    & Render \\
\hline
\textbf{Lego Knights} (1024x1024x5x5)      & 9.2 s (184) &  2.2 s                 &   35 ms   \\
\hline
\textbf{Lego Knights} (1024x1024x17x17)    & 56 s (403)  &  7.2 s                 &   35 ms   \\
\hline
\textbf{Chess Pieces} (1920x1080x5x5)      & 5.8 s (90)  &  4.3 s                 &   53 ms   \\
\hline
\textbf{Fruits} (625x434x9x9)              & 5.1 s (70)  &  0.9 s                 &   12 ms   \\
\hline
\textbf{Figurines} (625x434x9x9)           & 6.2 s (91)  &  0.9 s                 &   11 ms   \\
\hline
\textbf{Totoro WaterFall} (379x379x7x7)    & 3.8 s (72)  &  0.4 s                 &   7 ms    \\
\hline
\textbf{papillon} (768x768x9x9)            & 7.8 s (109) &  1.9 s                 &   20 ms   \\
\hline
\textbf{budha} (768x768x9x9)               & 8.1 s (113) &  1.9 s                  &  19 ms   \\
\hline
\end{tabular}
\vspace{-5pt}
\end{table}



\section{Direct Applications}
\label{sec:Applications}

\subsection{View Interpolation and Extrapolation}
\label{ssec:ViewInterpolation}

\begin{figure}[b]
\centering
\begin{minipage}[h]{.2\linewidth}
\centerline{\includegraphics[width=\linewidth]{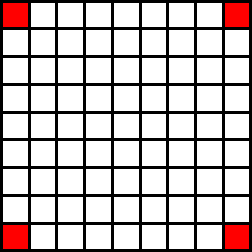}}
\vspace{-3pt}
\centerline{\small{(a)}}
\vspace{-5pt}
\end{minipage}
\hspace{.03\linewidth}
\begin{minipage}[h]{.2\linewidth}
\centerline{\includegraphics[width=\linewidth]{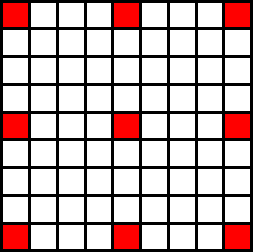}}
\vspace{-3pt}
\centerline{\small{(b)}}
\vspace{-5pt}
\end{minipage}
\hspace{.03\linewidth}
\begin{minipage}[h]{.2\linewidth}
\centerline{\includegraphics[width=\linewidth]{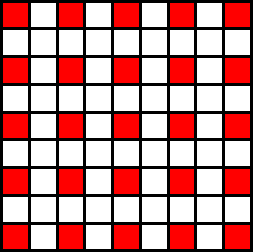}}
\vspace{-3pt}
\centerline{\small{(c)}}
\vspace{-5pt}
\end{minipage}
\hspace{.03\linewidth}
\begin{minipage}[h]{.2\linewidth}
\centerline{\includegraphics[width=\linewidth]{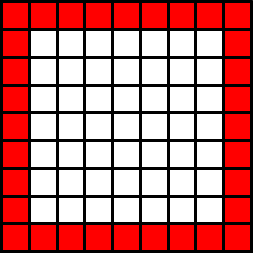}}
\vspace{-3pt}
\centerline{\small{(d)}}
\vspace{-5pt}
\end{minipage}
\caption{Tested sampling configurations of the input views (shown in red) for the view interpolation: (a) `2x2', (b) `3x3', (c) `5x5', (d) `border'.}
\label{fig:InterpConfigs}
\vspace{-6pt}
\end{figure}

\begin{figure*}
\centering
\begin{minipage}[h]{.24\linewidth}
\centerline{\includegraphics[width=\linewidth]{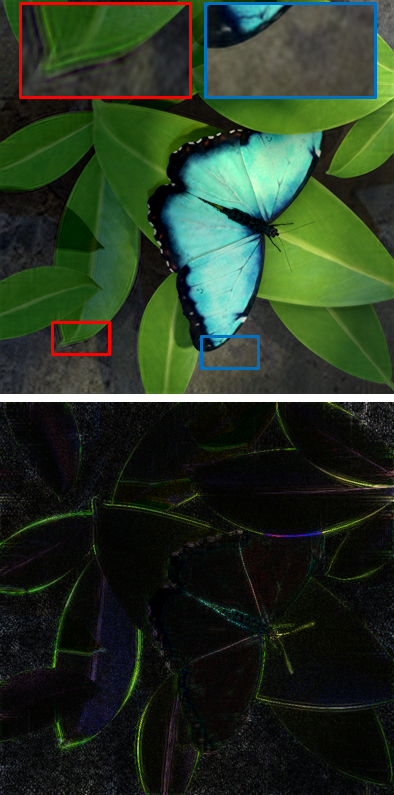}}
\vspace{-3pt}
\centerline{\small{(a)}}
\vspace{-3pt}
\end{minipage}
\begin{minipage}[h]{.24\linewidth}
\centerline{\includegraphics[width=\linewidth]{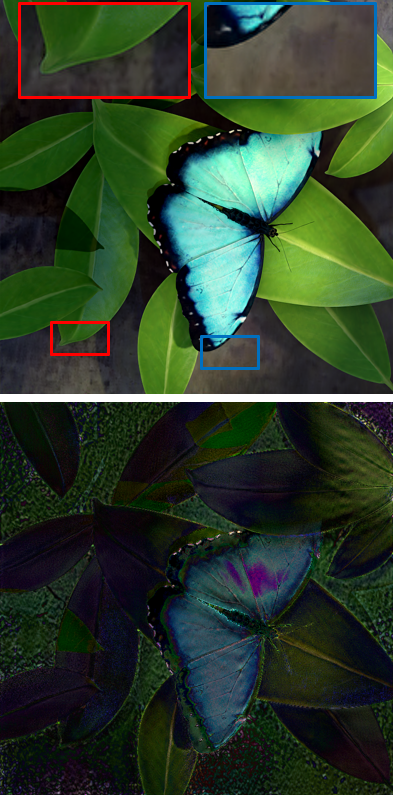}}
\vspace{-3pt}
\centerline{\small{(b)}}
\vspace{-3pt}
\end{minipage}
\begin{minipage}[h]{.24\linewidth}
\centerline{\includegraphics[width=\linewidth]{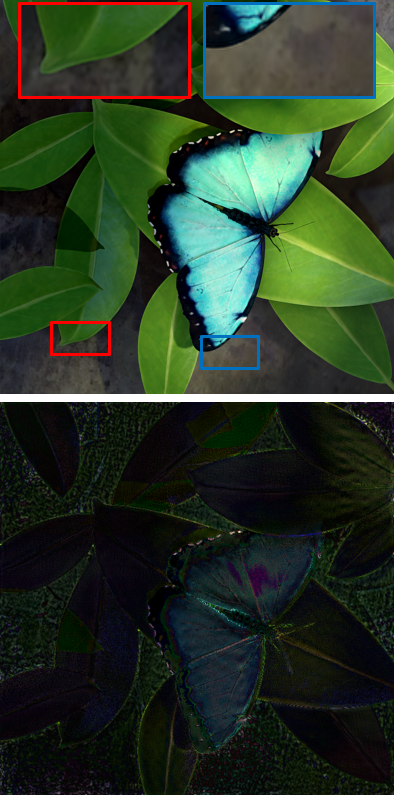}}
\vspace{-3pt}
\centerline{\small{(c)}}
\vspace{-3pt}
\end{minipage}
\begin{minipage}[h]{.24\linewidth}
\centerline{\includegraphics[width=\linewidth]{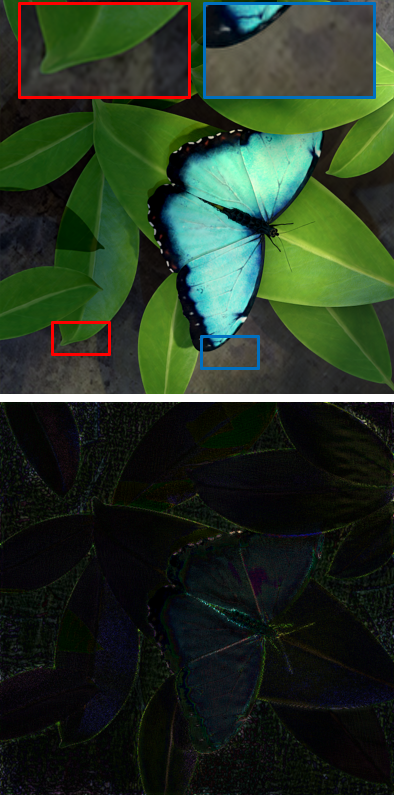}}
\vspace{-3pt}
\centerline{\small{(d)}}
\vspace{-3pt}
\end{minipage}
\caption{Interpolation of the central view from the 4 corner views of the Light Field `papillon' (i.e. 2x2 configuration) using: (a) only FDL, (b) only Shearlet \cite{Vaghar2018}, (c) `Shearlet(full)+FDL', (d) `Shearlet(border)+FDL'. The bottom images show the residual computed with the ground truth central view. A video of the results is available in the supplementary materials.}
\label{fig:InterpResult}
\vspace{-6pt}
\end{figure*}

As shown in Fig. \ref{fig:OursVsShiftAdd}, our approach reduces the angular aliasing and allows extending the aperture size for an input Light Field consisting of a sparse set of sub-aperture images. Equivalently, the layer construction can be seen as a view interpolation and extrapolation method since sub-aperture images with arbitrary angular coordinates can be rendered from the FDL model.

For evaluating the view interpolation capability of the method, we have compared our results with the recent method of Vagharshakyan et al. \cite{Vaghar2018} that enforces the sparsity of the epipolar images of the Light Field in the shearlet domain. For the comparison, the synthetic Light Field `papillon' was used with several sampling configurations of the input views as illustrated in Fig. \ref{fig:InterpConfigs}. The average PSNR of the interpolated views and the computation times are shown in Table \ref{tab:ViewInterp}. The table also includes the results for two schemes combining the FDL and the Shearlet approaches. In the `Shearlet(full)+FDL', the full Light Field is reconstructed with \cite{Vaghar2018}. Then, a FDL model is computed from the input views and the previously reconstructed intermediate views. The final reconstruction is obtained with FDL rendering. The `Shearlet(border)+FDL' scheme is similar, but only the views at the periphery of the Light Field are interpolated using \cite{Vaghar2018}. Our experiments with the Shearlet method were performed using the author's implementation of the epipolar image reconstruction with 100 iterations. The reconstruction of the full Light Field was performed by scanning the epipolar images in the `direct' order as detailed in \cite{Vaghar2018}.

\begin{table}[t]
\centering
\caption{View Interpolation results for the Light Field `papillon'. The results given are the average PSNR of the interpolated views and the total reconstruction time.}
\vspace{-5pt}
\label{tab:ViewInterp}
\begin{tabular}{|l|>{\centering\arraybackslash}m{.95cm}|>{\centering\arraybackslash}m{.95cm}|>{\centering\arraybackslash}m{.95cm}|>{\centering\arraybackslash}m{.95cm}|}
\hline
Method \;\textbackslash\, Input views   &   2x2                 &   3x3                     &   5x5                     &  border \\
\hline
\textbf{Shearlet} \cite{Vaghar2018} & 34.8 dB 8600 s            & 40.8 dB 9400 s            & 41.8 dB 11100 s           & 37.2 dB 6300 s \\
\hline
\textbf{FDL}                        & 36.8 dB \textbf{5.4 s}    & 40.9 dB \textbf{7.8 s}    & \textbf{42.9 dB 8.9 s}    & \textbf{43.1 dB 8.8 s} \\
\hline
\textbf{Shearlet(full)+FDL}         & 35.5 dB   8600 s          & 41.4 dB 9400 s            & 42.6 dB 11100 s           & 40.1 dB 6300 s \\
\hline
\textbf{Shearlet(border)+FDL}       & \textbf{38.4 dB} 3200 s   & \textbf{42.0 dB} 3200 s   & 42.7 dB 3200 s            & x \\
\hline
\end{tabular}
\vspace{-15pt}
\end{table}

The results in Table \ref{tab:ViewInterp} show that our FDL view interpolation performs particularly well in the `border' configuration.
The Shearlet method, on the other hand, does not fully take advantage of this configuration as it processes each epipolar image independently using only the input views located on the corresponding row or column in the Light Field.
In the other configurations, a higher PSNR is also observed with the FDL method. However, when the input Light Field is too sparse (e.g. 2x2), it produces ringing artifacts as shown in Fig. \ref{fig:InterpResult}. In this situation, the Shearlet method better preserves the edges than our FDL method, but it tends to blur the fine details and introduce color distortions, which explains the lower PSNR.
The best strategy for such sparse inputs is then to combine the two methods  with the `Shearlet(border)+FDL' scheme which does not produce ringing artifacts. In comparison with the Shearlet method alone, this scheme better keeps the details and color consistency as shown in Fig. \ref{fig:InterpResult}. It is also significantly faster since the reconstruction of the interior views is performed using our FDL approach which requires a negligible computing time compared to the Shearlet method.



\subsection{Denoising}
\label{ssec:Denoising}
Another direct application of our FDL construction algorithm is the Light Field denoising problem. Denoising is naturally obtained by constructing a FDL model from a Light Field, and by rendering images using the same angular coordinates as the input sub-aperture images. The noise is filtered in this approach thanks to the model definition that enforces the Light Field dimensionality gap prior \cite{Levin2010} (see Section \ref{ssec:Prior}).
Note that for the same reason, the method also ensures color and illumination consistency along the angular dimensions. Hence it can serve simultaneously as a denoising and color correction method for Light Fields with variations of color and illumination between the sub-aperture images. In practice, this is particularly useful for Light Fields captured with plenoptic cameras (e.g. Lytro), as shown in figure \ref{fig:DenoiseColor}. Additional results are also presented in the supplementary video.

\begin{figure}
\centering
\begin{minipage}[h]{.49\linewidth}
\centerline{\includegraphics[width=\linewidth]{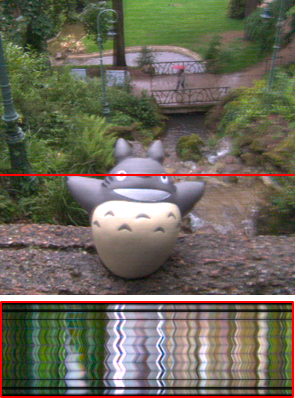}}
\vspace{-3pt}
\centerline{\small{(a)}}
\vspace{-3pt}
\end{minipage}
\begin{minipage}[h]{.49\linewidth}
\centerline{\includegraphics[width=\linewidth]{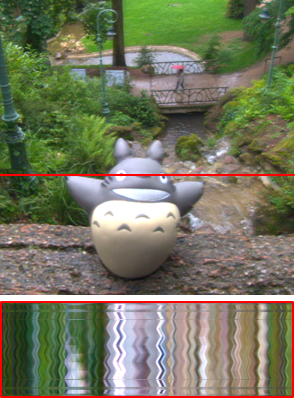}}
\vspace{-3pt}
\centerline{\small{(b)}}
\vspace{-3pt}
\end{minipage}
\caption{Simultaneous denoising and color correction of the Light Field `TotoroWaterfall' (captured with a Lytro camera): (a) Original (b) Processed with the FDL method. The top image corresponds to a view on the side of the Light Field. Epipolar images corresponding to the horizontal red line are shown on the bottom to visualize the color consistency between views.}
\label{fig:DenoiseColor}
\vspace{-6pt}
\end{figure}

However, for Light Fields with a wider baseline, our denoising may also introduce distortions due to the assumptions of non-occluded Lambertian Light Field in the problem definition. In order to better cope with this type of data, we propose a relaxation of the FDL model. In the original FDL calibration (see Section \ref{ssec:Calibration}), the shift applied to the $k^{th}$ layer to approximate the $j^{th}$ input sub-aperture image is given by the entry $\matr{P}_{j,k}$ of the matrix $\matr{P}=\vect{ud}^\top$. Since $\vect{u}$ and $\vect{d}$ are column vectors, the rank of the shift parameter matrix $\matr{P}$ is equal to 1. Hence, the shift applied to each layer is proportional to the angular coordinates of the view to reconstruct. However, this property is not suitable for non-lambertian surfaces and occlusions. Therefore, in the relaxed FDL model, we remove the rank 1 constraint by searching directly for the matrix $\matr{P}$ instead of $\vect{u}$ and $\vect{d}$ in the calibration algorithm. For that purpose, the gradient descent is applied using the gradients $\nabla\matr{P}$ defined in Eq. \eqref{eq:gradP}. In order to obtain satisfying convergence, the matrix $\matr{P}$ is initialized by the solution of the original rank 1 constrained problem. Once the matrix $\matr{P}$ is known, the denoised result is obtained by constructing the layers and rendering the sub-aperture images using the shifts $\matr{P}_{j,k}$ instead of $u_j\cdot d_k$.
Note that, for the denoising application, we only need to reconstruct the input sub-aperture images for which the shift parameters were computed.
Applying the relaxed model to the view interpolation application would also require interpolating the shift parameters for other view positions, which we do not consider in this paper.

\begin{figure}
\centerline{\includegraphics[width=\linewidth]{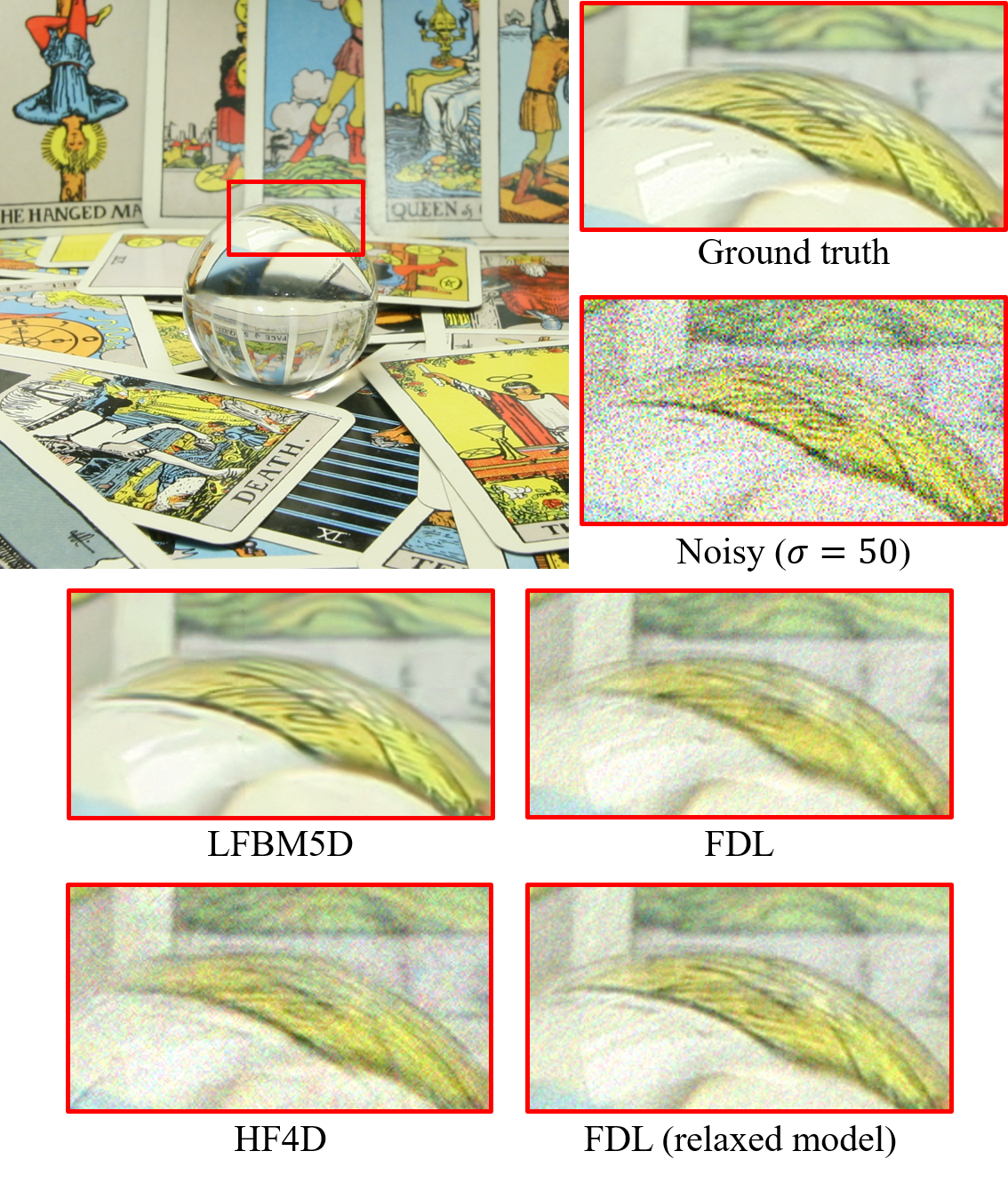}}
\vspace{-6pt}
\caption{Denoising results for the Light Field `Tarot'. A view on the corner of the Light Field is shown. For the test, the Light Field was corrupted by gaussian noise with standard deviation $\sigma=50$.}
\label{fig:DenoiseTarot}
\vspace{-6pt}
\end{figure}

Denoising results are presented in Fig. \ref{fig:DenoiseTarot} for the Light Field `Tarot' that exhibits strong non-lambertian effects. The reflections in the crystal ball are better preserved by the relaxed FDL model than the original one. For the comparison, the results obtained with the LFBM5D \cite{Alain2017} and the Hyperfan 4D filter (HF4D) \cite{Dansereau2013} are also shown. The best results were obtained with the LFBM5D method that completely removes the noise while preserving most of the details in the image. However, this method performs heavy processing and typically requires several hours for a Light Field.

The HF4D method is faster since it simply consists in multiplying the 4D Fourier Transform of the Light Field by a 4D filter. This hyperfan filter is designed to attenuate the frequencies outside of the theoretical region of support of the Light Field in the 4D Fourier domain, under the assumption of a non-occluded lambertian scene. With comparable processing times, our FDL approach removes more noise thanks to the linear optimization used instead of the direct filtering in the HF4D method. Furthermore, our model can be more easily generalized for the case of scenes with occlusions or non-lambertian surfaces, as shown with the proposed relaxation.



\section{Conclusion and Perpectives}
\label{sec:Conclusion}

We have presented a new representation for Light Fields called Fourier Disparity Layers. The light information in the scene is decomposed into several layers, each corresponding to a depth plane parameterized by a disparity value. The decomposition step is formulated in the Fourier domain using a simple linear least square problem per-frequency, hence allowing fast processing with GPU parallelization.
We have demonstrated the advantages of the FDL representation for several Light Field processing tasks. Those include real time rendering, view interpolation, denoising, as well as a calibration step that determines the angular coordinates of the input sub-aperture images along with an optimal set of disparity values for each layer.

The computational efficiency of our layer construction method coupled with its flexibility regarding the type of input images opens perspectives for an even larger range of applications. For example, our supplementary video presents results of FDL model construction from a focal stack, where the camera aperture and the focus parameter of each image are known. More generally, the method could even take advantage of a hybrid capture system taking images with different apertures, focusing depth, and varying positions on the camera plane. Furthermore, since our approach does not require a specific pattern of the input view positions, it can greatly simplify the capture of Light Fields by removing the need for the different viewpoints to be regularly sampled on a grid.

However, despite this flexibility, we have shown that the quality of the FDL model produced with our method depends on the input configuration. For example, while very accurate view interpolation is obtained from the sub-aperture images located on the borders of the Light Field, a too sparse viewpoint sampling may lead to ringing artifacts.
Furthermore, for Light Fields with a large baseline, the FDL model is more likely to produce visible errors due to large occlusion areas or non-lambertian effects.
In the paper, we have partially addressed these limitations. In the case of a too sparse viewpoint sampling, a combination of our view interpolation with a state of the art approach was shown to outperform either of the two methods taken individually. A better handling of occlusions and non-lambertian effects was also obtained, for the denoising application, using a relaxed version of the model.

In the aim of further extending the applicability of the FDL approach, future work may focus on generalizing the relaxed model to view interpolation, or including an additional prior directly in the layer construction (e.g. sparsity in the Shearlet domain similarly to \cite{Vaghar2018}) to better cope with very sparse or very noisy Light Fields. A generalization of the calibration to wide aperture images would also be a valuable tool to facilitate the creation of Light Fields from less conventional input data such as focal stacks.

\begin{appendices}

\section{Proof of sparsity prior}
\label{app:prior}


The spatial regions $\Omega_k$ are defined for the central view at $u=0$. The corresponding regions $\Omega_k^{u}$ can also be defined for any view $u$ by \mbox{$\Omega_k^{u}=\left\{x\in\mathbb{R}\mid x+ud_k\in\Omega_k\right\}$}.

The expression in Eq. \eqref{eq:Prior1} is then formulated as:
\begin{equation}
\label{eq:Prior2}
\forall k\in \ldbrack 1,n\rdbrack, \forall (x,u) \in \Omega_k^{u}\times\mathbb{R}, L(x,u) = L(x+ud_k,0).
\end{equation}

In the assumption of a non-occluded Light Field, the regions $\Omega_k^{u}$ are such that \mbox{$\forall u\in\mathbb{R}, x\in\mathbb{R}\setminus\bigcup\limits_{k}\Omega_k^{u}$, $L(x,u)=0$}, and for any fixed $u\in\mathbb{R}$, the sets $\Omega_k^{u}$ are pairwise disjoint. Hence, the Fourier Transform of the Light Field is given by:
\begin{equation}
\begin{split}
&\hat{L}(\omega_x,\omega_u) = \iint_{\mathbb{R}^2} e^{-2i\pi(x\omega_x+u\omega_u)} L(x,u)\mathrm{d}x\mathrm{d}u \\
& = \quad \int_{-\infty}^{+\infty} e^{-2i\pi u\omega_u}\sum_k \left[\int_{\Omega_k^u} e^{-2i\pi x\omega_x} L(x,u) \mathrm{d}x\right]\mathrm{d}u.
\end{split}
\end{equation}
Using Eq. \eqref{eq:Prior2}, and by change of variable we obtain:
\begin{equation}
\begin{split}
&\hat{L}(\omega_x,\omega_u) \\
& = \int_{-\infty}^{+\infty} e^{-2i\pi u\omega_u}\sum_k \left[\int_{\Omega_k^u} e^{-2i\pi x\omega_x} L(x+ud_k,0) \mathrm{d}x\right]\mathrm{d}u \\
& = \int_{-\infty}^{+\infty} e^{-2i\pi u\omega_u}\sum_k \left[\int_{\Omega_k} e^{-2i\pi (x-ud_k)\omega_x} L(x,0) \mathrm{d}x\right]\mathrm{d}u.
\end{split}
\end{equation}
By re-arranging the terms the result is:
\begin{equation}
\begin{split}
&\hat{L}(\omega_x,\omega_u) \\
& = \sum_k \left[\int_{-\infty}^{+\infty} e^{-2i\pi u(\omega_u-d_k\omega_x)}\mathrm{d}u  \int_{\Omega_k} e^{-2i\pi x\omega_x} L(x,0) \mathrm{d}x \right] \\
& = \sum_k \left[\delta(\omega_u-d_k\omega_x) \hat{L}^k(\omega_x) \right].
\end{split}
\end{equation}
where we define $\hat{L}^k(\omega_x)=\int_{\Omega_k} e^{-2i\pi x\omega_x} L(x,0) \mathrm{d}x$. \hfill $\square$

\end{appendices}

\bibliographystyle{IEEEtran}
\bibliography{references}

\end{document}